\documentclass{article} 
\usepackage{nips12submit_e,times}
\usepackage{amssymb}
\usepackage{amsmath}
\usepackage{subfig}
\usepackage{amsthm}
\usepackage{graphicx}
\usepackage{cite}
\usepackage{algorithm}
\usepackage{algpseudocode}

\title{Hierarchical Data Representation Model - Multi-layer NMF}

\author{
Hyun Ah Song\\
Department of Electrical Engineering\\
KAIST\\
Daejeon, 305-701 \\
\texttt{hyunahsong@kaist.ac.kr} \\
\And
Soo-Young Lee \\
Department of Electrical Engineering \\
KAIST\\
Daejeon, 305-701 \\
\texttt{sylee@kaist.ac.kr} \\
}

\nipsfinalcopy 

\begin{document}

\maketitle
\begin{abstract}
In this paper, we propose a data representation model that demonstrates hierarchical feature learning using nsNMF. We extend unit algorithm into several layers. Experiments with document and image data successfully discovered feature hierarchies. 
We also prove that proposed method results in much better classification and reconstruction performance, especially for small number of features.
\end{abstract}


\section{Introduction}
\label{introduction}
In order to understand complex data, hierarchical feature extraction strategy has been used [1]. One best known algorithm is Deep Belief Network (DBN) introduced in 2006 [2]. With the success of training deep architectures, several variants of deep learning have been introduced [3]. Although these multi-layered algorithms take hierarchical approaches in feature extraction and provide efficient solution to complex problems, they do not provide us the relationships of features in form of hierarchies that are learned throughout the hierarchical structure.

In this paper, we propose a hierarchical data representation model, hierarchical multi-layer non-negative matrix factorization. (Similar approach has been introduced in [4].) We extend a variant of NMF algorithm [5], nsNMF [6] into several layers for hierarchical learning. Here, we demonstrate intuitive feature hierarchies present in the data set by learning relationships between features across layers. We also prove that instead of one step learning, hierarchical approach learns more meaningful and helpful features, which leads to better distributed representations, and results in better performance in classification and reconstruction for small number of features, which guarantees reduced loss of performance, even when representing data in small dimensions.


\section{Non-smooth non-negative matrix factorization (nsNMF)}
Proposed network is constructed by stacking nsNMF [6] into several layers.
Non-smooth non-negative matrix factorization (nsNMF) is a variant of NMF that restricts sparsity constraint. Basic NMF decomposes non-negative input data \textbf{X} into non-negative \textbf{W} and \textbf{H}, which are features and corresponding coefficients or data representation respectively. It aims to reduce error between original data \textbf{X} and its reconstruction \textbf{WH}: $C = \frac{1}{2} \|\mathbf{X}-\mathbf{WH}\|^2 = \frac{1}{2} \sum_{i=1}^m\sum_{j=1}^n(X_{ij}-\sum_{k=1}^l W_{ik}H_{kj})^2.$

%


To apply sparsity constraint to standard NMF, a sparsity matrix \textbf{S} is introduced in [6]: $ \mathbf{S} = (1-\theta)\mathbf{I}(k) + \frac{\theta}{k}\mathbf{ones}(k).$
$k$ is number of features, and $\theta$ is parameter for smoothing effect, in range of 0 to 1. \textbf{I}(k) is identity matrix of size k x k, and \textbf{ones}(k) is a matrix of size k x k with all components of 1s. We smooth a matrix by multiplying it with \textbf{S}. The closer $\theta$ is to 1, more smoothing effect is applied. During alternative update, we smooth \textbf{H} matrix by multiplying \textbf{S} and \textbf{H} during iterations as \textbf{H=SH}. To compensate the loss of sparsity, \textbf{W} becomes sparse.

\section{Multi-layer architecture}
\label{Multi-layer architecture}
The proposed hierarchical multi-layer NMF structure comprise of several layers of unit algorithm. 
We first train each layer separately.
We process outcome of each layer $\mathbf{H}^{(l)}$ to get $\mathbf{K}^{(l)}$. $K_{kj}^{(l)}=f\left(\frac{H_{kj}^{(l)}}{M_{kj}^{(l)}}\right)$, where $M_{kj}^{(l)}= \sum_{j'=1}^n \frac{ H_{kj'}^{(l)}}{n}$, f($\cdot$) is nonlinear function, and $l$ denotes index of layer, $l = 1, 2, ... L$. The superscript of each term denotes layer index. 
Processed data representation of $\mathbf{K}^{(l)}$ is used as input to next layer. Using nsNMF, $\mathbf{K}^{(l)}$ is decomposed into $\mathbf{W}^{(l+1)}$ and $\mathbf{H}^{(l+1)}$:$\mathbf{K}^{(l)}\approx\mathbf{W}^{(l+1)}\mathbf{H}^{(l+1)}$. 
Then, we use outcome of separate training as initialization, and train the whole network jointly. The cost function for joint training is described: $C = \frac{1}{2} \sum_{i=1}^m\sum_{j=1}^n(X_{ij}-\sum_{k=1}^l W_{ik}^{(1)}\widetilde{H_{kj}^{(1)}})^2$, 
where $\widetilde{H_{kj}^{(l)}}$ is the reconstruction of $H_{kj}^{(l)}$, which can be computed via back propagation of errors from the last layer to the $l^{th}$ layer:
$\widetilde{\mathbf{H}^{(L-1)}}\approx\mathbf{M}^{(L-1)}\odot f^{-1}\left(\mathbf{W}^{(L)}\widetilde{\mathbf{H}^{(L)}}\right)$,...,
$\widetilde{\mathbf{H}^{(1)}}\approx\mathbf{M}^{(1)}\odot f^{-1}\left(\mathbf{W}^{(2)}\widetilde{\mathbf{H}^{(2)}}\right)$, where $\widetilde{\mathbf{H}^{(L)}}=\mathbf{H}^{(L)}$. 
$f^{-1}(\cdot)$ is inverse nonlinear function. (more details on the actual update computation is described in Appendix \ref{AA}).
After training until the last layer, final data representation $\mathbf{H}^{(L)}$ is acquired. This is the activation information of complex features, which is the integration of features throughout the layers, $\mathbf{W}^{(1)}\mathbf{W}^{(2)}...\mathbf{W}^{(L)}$.

For more detailed explanation, refer to the pseudo-code for the training procedure in Appendix \ref{Pseudo-code}.

\section{Document data feature hierarchies}
\label{Document data feature hierarchies}

We applied our proposed network to document database. We used ''Reuters-21578 collection, distribution 1.0''\footnote{The Reuters-21578, Distribution 1.0 test collection is available from David D. Lewis’ professional home page, currently: http://www.research.att.com/~lewis} as database. We sorted top 10 categories from ModApte split, conducted pre-processing of removing stop-words, and reduced dimension to 1000. There are 5786 and 2587 document samples for training data, and test data. We constructed two-layered network with number of hidden neurons as 160.

\begin{figure}
\centering
\includegraphics[height=6cm]{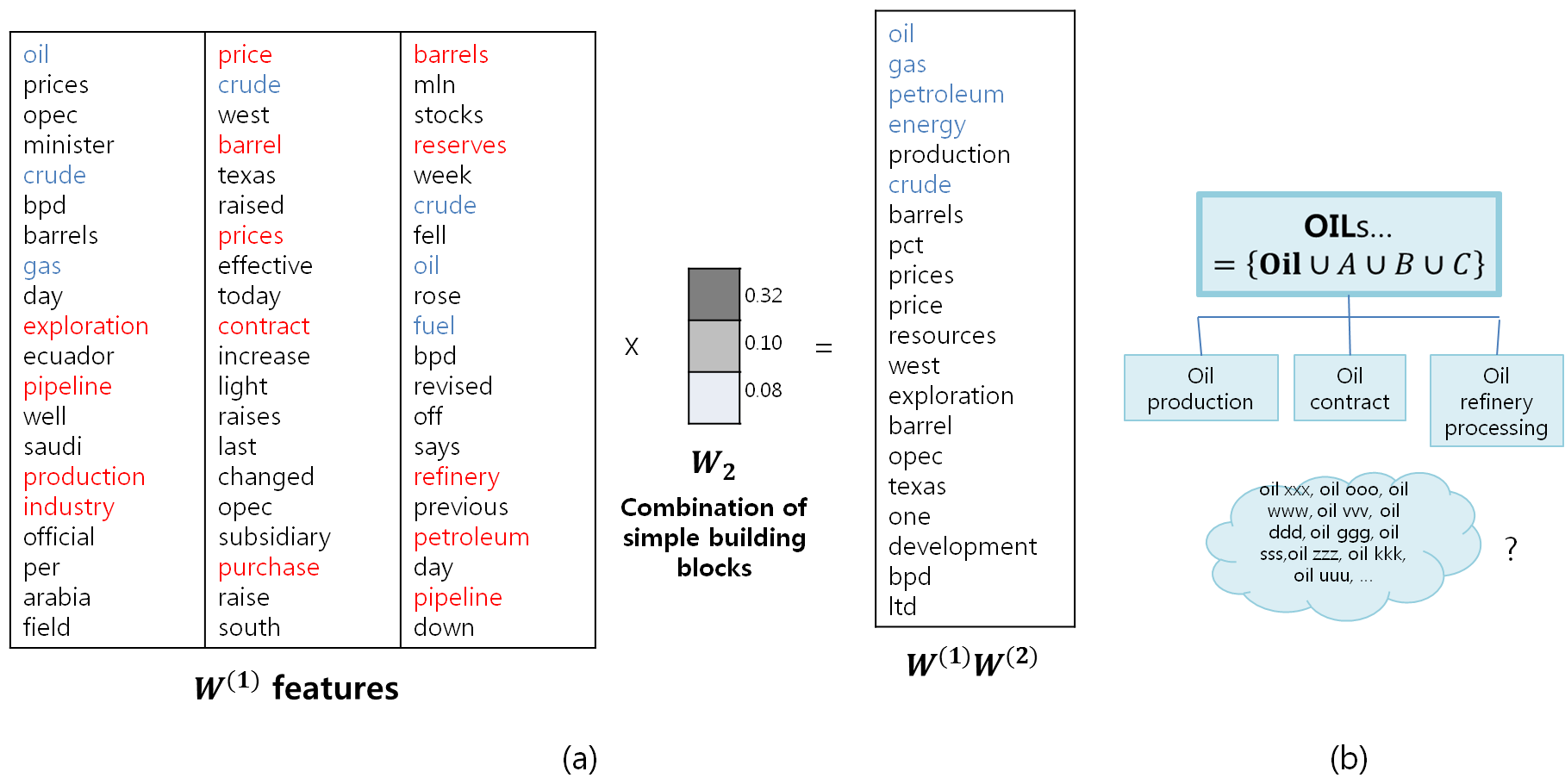}
\caption{Concept hierarchies in Reuters. (a) Experimental results, and (b) diagram of result in (a).}
\label{fig:8}
\end{figure}

\begin{figure}
\centering
\includegraphics[height=3.2cm]{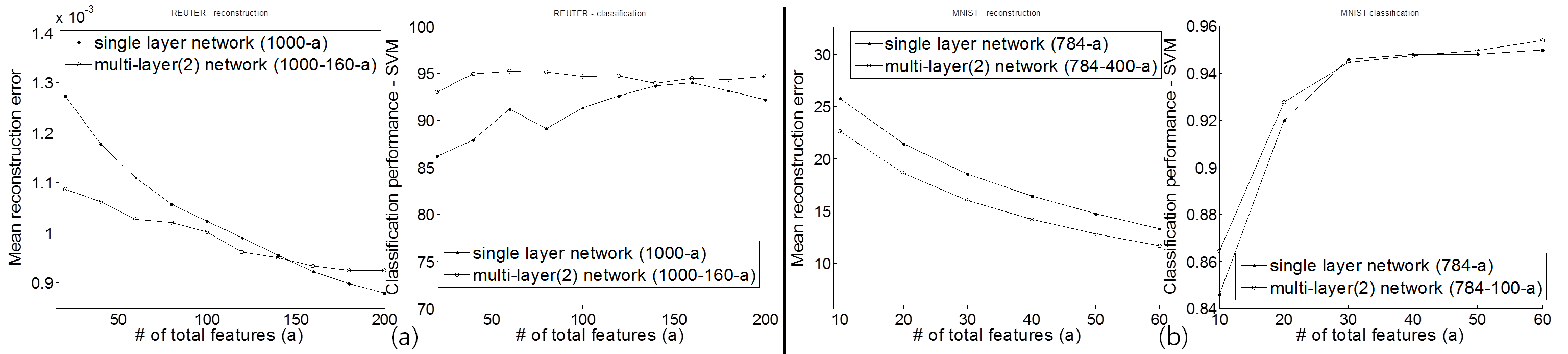}
\caption{Reconstruction error \textit{(left)} and classification rate \textit{(right)} of (a) Reuters and (b) MNIST.}
\label{fig:99}
\end{figure}

We observed how concepts form hierarchies in document data in Figure \ref{fig:8} (a). 
First, second, and third $\mathbf{W}^{(1)}$ features contain words related to 'oil production' (exploration, pipeline, production, industry), 'oil contract' (contract, purchase, barrel, prices), and  'oil refinery processing' (refinery, reserves, pipeline, petroleum), respectively. These sub-class topic features are combined together and develop into one broader topic 'oil.' 
With this combination relationship of features, we can figure out that those three seemingly independent features can be re-categorized under the same broader topic. (The concept hierarchy learned in Reuters: sub-categories of 'oil production', 'contract', and 'refinery processing' exist under 'oil' category.) 
Furthermore we analyzed reconstruction and classification performance as shown in Figure \ref{fig:99} (a). 
The proposed hierarchical feature extraction method results in much better classification and reconstruction, especially for small number of features, compared to extracting features at one step. This proves the efficiency and effectiveness of our proposed approach in learning of features.

\begin{figure}
\centering
\includegraphics[height=3.2cm]{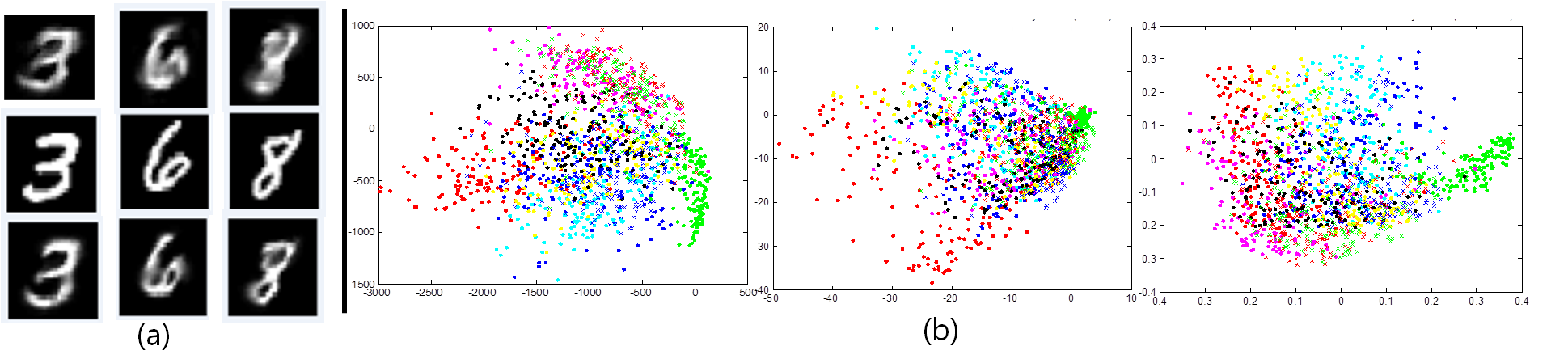}
\caption{(a) Reconstruction by shallow network (first row) and proposed network (third row), with original input (second row). (b) Final data representation (visualized by PCA) comparison of shallow network \textit{(middle)} and proposed network \textit{(right)}, in comparison to raw data \textit{(left)}}
\label{fig:10}
\end{figure}

We also applied our network to handwritten digit image MNIST\footnote{Available at: http://yann.lecun.com/exdb/mnist/}. The final data representation $\textbf{H}^L$ displayed distinct activation patterns for samples of the different classes, as a result of 
successful learning of feature hierarchy, which determines the combination of low level features in forming of distinct class features.
In Figure \ref{fig:99} (b), the reconstruction error and classification performance also demonstrate better performance of our proposed method in small number of dimensions.
In Figure \ref{fig:10} (a), we can observe sparser and clear reconstruction of our proposed network. 
The Fisher discriminant values of final data representation of the shallow network and our proposed network were 0.51 and 0.61 respectively. We can infer that proposed network learns more meaningful and helpful features so that it results in better distributed (clustered) representation of data.
We can also check this via the visualization of $\mathbf{H}^{(L)}$ to 2-D domain shown in Figure \ref{fig:10} (b).



\section{Conclusion}
\label{Conclusion}



In this paper, we proposed a hierarchical data representation model, hierarchical multi-layer NMF by stacking nsNMF into several layers. We demonstrated hierarchical approach in learning of the features. There are mainly two findings of our research. Taking hierarchical learning by stacking NMFs: 1.reveals intuitive feature hierarchies (subcategories) by learning feature relationships throughout the layers, and 2.learns more meaningful features compared to one-step learning. (as a result, our proposed method
results in much better classification and reconstruction performance, provided small number of dimensions for data representation.)
We expect our proposed method to be applied to various types of data for discovering underlying feature hierarchies and at the same time, maintain reconstruction and classification performance even with small number of features for data representation.

%

\subsubsection*{References}

\small{
[1] Bengio, Y. (2007). Learning Deep Architectures for AI. Foundations and Trends in Machine Learning, 2(1), 1-127.

[2] Hinton, G. E., Osindero, S., and Teh, Y. (2006). A fast learning algorithm for deep belief nets. Neural Computation, 18, 1527-1554.

[3] Bengio, Y., Lamblin, P., Popovici, D., and Larochelle, H. (2007). Greedy Layer-Wise Training of Deep Networks. NIPS,153-160.

[4] Ahn, J., Choi, S., Oh, J. (2004). A Multiplicative Up-Propagation Algorithm. Proceedings of the 21st International Conference on Machine Learning.

[5] Lee, D. D., and Seung, H. S. (1999). Learning the parts of objects by non-negative matrix factorization. Nature, 401, 788-791.

[6] Pascual-Montano, A., Carazo, J. M., Kochi, K., Lehmann, D., and Pascual-Marqui, R. D. (2006). Non-Smooth Nonnegative Matrix Factorization (nsNMF). IEEE Trans. Pattern Anal. Machine Intell, 403-415.

}

\newpage
\appendix
\section{Appendix: Actual update computation}
\label{AA}
Continued from Section \ref{Multi-layer architecture}, the actual computation is done as described in~\eqref{eq:update1}.
\begin{subequations}\label{eq:update1}
\begin{align}
   W^{(l)}_{ik} \leftarrow W^{(l)}_{ik}\frac{\left(\mathbf{Nu}^{(l)}\widetilde{\mathbf{H}^{(l)^{T}}}\right)_{ik}}{\left(\mathbf{De}^{(l)} \widetilde{\mathbf{H}^{(l)^{T}}}\right)_{ik}} \textit{, and }
      H^{(l)}_{kj} \leftarrow H^{(l)}_{kj}\frac{\left(\mathbf{W}^{(l)^{T}}\mathbf{Nu}^{(l)}\right)_{kj}}{\left(\mathbf{W}^{(l)^{T}}\mathbf{De}^{(l)}\right)_{kj}} \textit{, where }\\
\mathbf{Nu}^{(l)} = 
\begin{cases}
	\mathbf{X} & \textit{if l = 1}\\	\left(\mathbf{W}^{(l-1)^{T}}\mathbf{Nu}^{(l-1)}\right)\odot\left(\mathbf{M}^{(l-1)}f^{-1'}\left(\mathbf{W}^{(l)}\mathbf{H}^{(l)}\right)\right) & \textit{otherwise}
\end{cases} \\
\mathbf{De}^{(l)} = 
\begin{cases}
	\widetilde{\mathbf{X}} & \textit{if l = 1}\\	     \left(\mathbf{W}^{(l-1)^{T}}\mathbf{De}^{(l-1)}\right)\odot\left(\mathbf{M}^{(l-1)}f^{-1'}\left(\mathbf{W}^{(l)}\mathbf{H}^{(l)}\right)\right) & \textit{otherwise}
\end{cases}
\end{align}
\end{subequations}

Here, 
$\widetilde{\mathbf{X}} = \mathbf{W}^{(1)}\widetilde{\mathbf{H}^{(1)}}$.
$\widetilde{H_{kj}^{(l)}}$ is the reconstruction of $H_{kj}^{(l)}$, which can be computed via back propagation of errors from the last layer to the $l^{th}$ layer as shown in~\eqref{eq:Hrecon}.
\begin{equation}\label{eq:Hrecon}
\widetilde{\mathbf{H}^{(l)}}=
\begin{cases}
\mathbf{H}^{(l)}  & \textit{if $l = L$}\\
\mathbf{M}^{(l)}\odot f^{-1}\left(\mathbf{W}^{(l+1)}\widetilde{\mathbf{H}^{(l+1)}}\right) & \textit{if $l = L-1, ... , 1$}	
\end{cases}
\end{equation}
$\mathbf{M}^{(l)}$ is a matrix of column-wise mean of $\mathbf{H}^{(l)}$, and $f^{-1}(\cdot)$ is inverse nonlinear function.

\clearpage
\section{Appendix: Pseudo-code for training procedure of proposed network}
\label{Pseudo-code}

\begin{algorithmic}
\State \%\% Separate training of layers in extending mode
\For{$l=1:L$}
	\State Randomly initialize $\mathbf{W}^{(l)}$ and $\mathbf{H}^{(l)}$
	\If {$l=1$}
   		\State $\mathbf{K}^{(l-1)} = \mathbf{X}$
	\EndIf
	\For{$iteration=1:(until convergence)$}
		\State $W^{(l)}_{ik} \leftarrow W^{(l)}_{ik}\frac{(\mathbf{K}^{(l-1)}\mathbf{H}^{(l)^T})_{ik}}{(\mathbf{W}^{(l)}\mathbf{H}^{(l)}\mathbf{H}^{(l)^T})_{ik}}$
		\State
  		$H^{(l)}_{kj} \leftarrow H^{(l)}_{kj}\frac{(\mathbf{W}^{(l)^T}\mathbf{K}^{(l-1)})_{kj}}{(\mathbf{W}^{(l)^T}\mathbf{W}^{(l)}\mathbf{H}^{(l)})_{kj}}$
	\EndFor
	\State $M_{kj}^{(l)}= \sum_{j'=1}^n H_{kj'}^{(l)}/n$\;
	\State $K_{kj}^{(l)}=f\left(\frac{H_{kj}^{(l)}}{M_{kj}^{(l)}}\right)$\;		
\EndFor
\State
\State \%\% Joint training the whole network
\State Use $\mathbf{W}^{(l)}$ and $\mathbf{H}^{(l)}$, and use $\mathbf{M}^{(l)}$ acquired from above
\For{$iteration=1:(until convergence)$}	
	\For{$l=1:L$}
		\If{$l = L$}
			\State $\widetilde{\mathbf{H}^{(l)}}=\mathbf{H}^{(l)}$
		\Else
			\State $\widetilde{\mathbf{H}^{(l)}}=\mathbf{M}^{(l)}\odot f^{-1}\left(\mathbf{W}^{(l+1)}\widetilde{\mathbf{H}^{(l+1)}}\right)$	
			\State which can be written in full length as: 
			\State $\widetilde{\mathbf{H}^{(l)}} = \mathbf{M}^{(l)}\odot f^{-1}(\mathbf{W}^{(l+1)}(\mathbf{M}^{(l+1)}\odot f^{-1}(\mathbf{W}^{(l+2)}(...(\mathbf{M}^{(L-1)}\odot f^{-1}(\mathbf{W}^{(L)}\mathbf{H}^{(L)})))))$
		\EndIf
		\If{$l=1$}
			\State $\mathbf{Nu}^{(l)} = \mathbf{X}$
			\State $\mathbf{De}^{(l)} = \widetilde{\mathbf{X}}$ 
			\State which can be written in full length as:
			\State $\widetilde{\mathbf{X}} = \mathbf{W}^{(1)}\widetilde{\mathbf{H}^{(1)}} = \mathbf{W}^{(1)}(\mathbf{M}^{(1)}\odot f^{-1}(\mathbf{W}^{(2)}(\mathbf{M}^{(2)}\odot f^{-1}(\mathbf{W}^{(3)}( ...$
			\State $\mathbf{M}^{(L-1)}\odot f^{-1}(\mathbf{W}^{(L)}\mathbf{H}^{(L)}))))))$ 
		\Else
			\State $\mathbf{Nu}^{(l)} = \left(\mathbf{W}^{(l-1)^{T}}\mathbf{Nu}^{(l-1)}\right)\odot\left(\mathbf{M}^{(l-1)}f^{-1'}\left(\mathbf{W}^{(l)}\mathbf{H}^{(l)}\right)\right)$
			\State $\mathbf{De}^{(l)} = \left(\mathbf{W}^{(l-1)^{T}}\mathbf{De}^{(l-1)}\right)\odot\left(\mathbf{M}^{(l-1)}f^{-1'}\left(\mathbf{W}^{(l)}\mathbf{H}^{(l)}\right)\right)$
		\EndIf
		\State $W^{(l)}_{ik} \leftarrow W^{(l)}_{ik}\frac{\left(\mathbf{Nu}^{(l)}\widetilde{\mathbf{H}^{(l)^{T}}}\right)_{ik}}{\left(\mathbf{De}^{(l)} \widetilde{\mathbf{H}^{(l)^{T}}}\right)_{ik}}$
		\State $ H^{(l)}_{kj} \leftarrow H^{(l)}_{kj}\frac{\left(\mathbf{W}^{(l)^{T}}\mathbf{Nu}^{(l)}\right)_{kj}}{\left(\mathbf{W}^{(l)^{T}}\mathbf{De}^{(l)}\right)_{kj}}$
	\EndFor
\EndFor
\end{algorithmic}

\clearpage

\end{document}